\documentclass[a4paper]{article}
\usepackage{spconf}

\usepackage[utf8]{inputenc}
\usepackage{amssymb,amsmath,bm,graphicx}
\graphicspath{{figures/}}

\usepackage{textcomp}

\ninept

\title{Optimising The Input Window Alignment in CD-DNN Based Phoneme Recognition for Low Latency Processing}
\name{Akash Kumar Dhaka and Giampiero Salvi}
\address{KTH Royal Institute of Technology,\\
School of Computer Science and Communication,\\
Dept. for Speech, Music and Hearing,
Stockholm, Sweden\\
\texttt{\{akashd, giampi\}@kth.se}}
\begin{document}
%
\maketitle
\begin{abstract}
We present a systematic analysis on the performance of a phonetic recogniser when the window of input features is not symmetric with respect to the current frame.
The recogniser is based on Context Dependent Deep Neural Networks (CD-DNNs) and Hidden Markov Models (HMMs).
The objective is to reduce the latency of the system by reducing the number of future feature frames required to estimate the current output.

Our tests performed on the TIMIT database show that the performance does not degrade when the input window is shifted up to 5 frames in the past compared to common practice (no future frame).
This corresponds to improving the latency by 50 ms in our settings.
Our tests also show that the best results are not obtained with the symmetric window commonly employed, but with an asymmetric window with eight past and two future context frames, although this observation should be confirmed on other data sets.

The reduction in latency suggested by our results is critical for specific applications such as real-time lip synchronisation for tele-presence, but may also be beneficial in general applications to improve the lag in human-machine spoken interaction.

\end{abstract}
%
%
\vspace{-2mm}
\section{Introduction}
\label{sec:intro}
In recent years, the development of deep neural models based on Restricted Boltzman machines (RBMs) pretraining has revitalised the use of artificial neural networks (ANNs) in automatic speech recognition (ASR) as well as in many other fields (see \cite{LeCunEtAl2015Nature, Schmidhuber2015NeuralNetworks} for extensive reviews).
A key factor that determines the usability of applications based on speech recognition is the latency or lag of the system.
In dialogue systems, e.g., long latencies may disrupt the natural turn-taking in the human-machine conversation.
In other specific applications the lag may even be more critical.
A typical example involves systems that use ASR to drive the lip movements of an avatar in real time to support telepresence \cite{gs:SalviEtAl2009, MuEtAl2010LipSync, LiEtAl2013ICASSP}.

The latency in a typical speech recogniser based on a hybrid between Neural Networks (NNs) and Hidden Markov Models (HMMs) is determined by a number of factors:
\begin{itemize}
\item the hardware (sound card) introduces some lag in digitising the speech samples and making them available to the drivers. Typical values are in the order of milliseconds;
\item the speech samples are returned by the driver in buffers of a certain size (this could be as long as half a second, but can be reduced to a few ms);
\item in spectral based feature extraction, speech samples are grouped into windows (frames) often around 25-40 ms in length;
\item many methods for feature extraction also compute time derivatives of the features, which require a number of frames in the past and the future. These often include three context frames for the first and three for the second derivatives for a total of 60 ms, if we assume 10 ms spaced feature vectors;
\item the input to the neural network that estimates state probabilities may include a window of context frames. Typical values are 5 future and 5 past frames that correspond to 50 ms latency. In some cases the context may extend to the whole utterance, e.g., in some application of convolutional neural networks;
\item the decoder that combines the probability estimates produced by the NN with the HMM time model usually requires a certain look-ahead (from a few hundreds of ms to the whole utterance).
\end{itemize}

In \cite{gs:Salvi2006} we used a hybrid of Recurrent Neural Networks (RNNs) and HMMs that was specifically designed for low latency processing.
The feature extraction was based on Mel Frequency Cepstral Coefficients (MFCCs) without time derivatives and the RNN did not receive any future feature frame in input, thus limiting the latency of the RNN to the size of the feature extraction window.
The purpose of that study, however, was limited to evaluating the effect of varying the look-ahead length of the Viterbi decoder in the system.

Nearly all recent methods that use DNN based acoustic models for ASR employ symmetric context windows as input \cite{Mohamed12, pdnn, dbn09, YaoEtAl2012SLTadaptation, DahlEtAl2011ICASSP} and are therefore affected by a certain latency.
In \cite{PeddintiEtAl2015}, time delayed neural networks were used with asymmetric context windows in some cases. Similarly in \cite{LeiEtAl2013} context windows with more frames on the left context were used. However, we are not aware of a systematic and detailed investigation on the effect of the context window asymmetry.

In this study, we want to determine the relationship between latency of the DNN model and its performance.
In order to do this, we analyse how the performance of the recogniser proposed in \cite{pdnn} varies as the alignment of the input context window is shifted back or forward in time. It is not our intent to report on state-of-the-art results, but to give and indication on the relative effects of shifting the input window. Also, we report results on Phoneme Error Rate (PER) on the TIMIT data, because we want to have precise control over shifts in time and therefore require carefully annotated data. 



\section{METHOD}
\label{sec:method}
For our experiments, we use a Context Dependent Deep Neural Network (CD-DNN) trained to estimate the posterior probabilities of a set of senones given a sequence of input feature vectors. The probability estimations are then used in a Hidden Markov Model (HMM) in combination with a bigram phoneme-level language model for phonetic recognition.

The set of senones and their alignment with the speech utterances in the dataset is determined by training a context dependent HMM recogniser based on Gaussian Mixture Models.
The number of senones is reduced with decision tree based clustering.

The DNN training procedure has been well described in \cite{dbn09}.
The weights in the CD-DNN are initialised using a Deep Belief Network (DBN), that is, a stack of Restricted Boltzmann Machines (RBMs).
The DBN is trained generatively by fitting the layers one at a time (greedily) by means of the contrastive divergence procedure.
The final output layer in the CD-DNN is a generalised softmax (GSM) layer representing a distribution over senones.
Given the generative initialisation, the full model is fine-tuned with back-propagation training.
The input to the model is a context window of $n$ successive frames of raw filterbank feature vectors.
The feature vectors are normalised to zero mean and unit variance. 
We use filterbank features instead of MFCCs because they have been reported to achieve good results in combination with DNNs without the need for time derivatives that would increase the latency \cite{Jaitly2014Thesis}.

We use a Viterbi decoder to generate the phoneme sequences and a phoneme-level bigram model estimated on the training set.
The acoustic scale used in the decoder to tune acoustic and language models was optimised for each test independently on the development test and the optimal value was then used on the test set.
The decoder follows the lattice generation and pruning approach described in \cite{danp12}.

\begin{figure}
\includegraphics[width=\columnwidth]{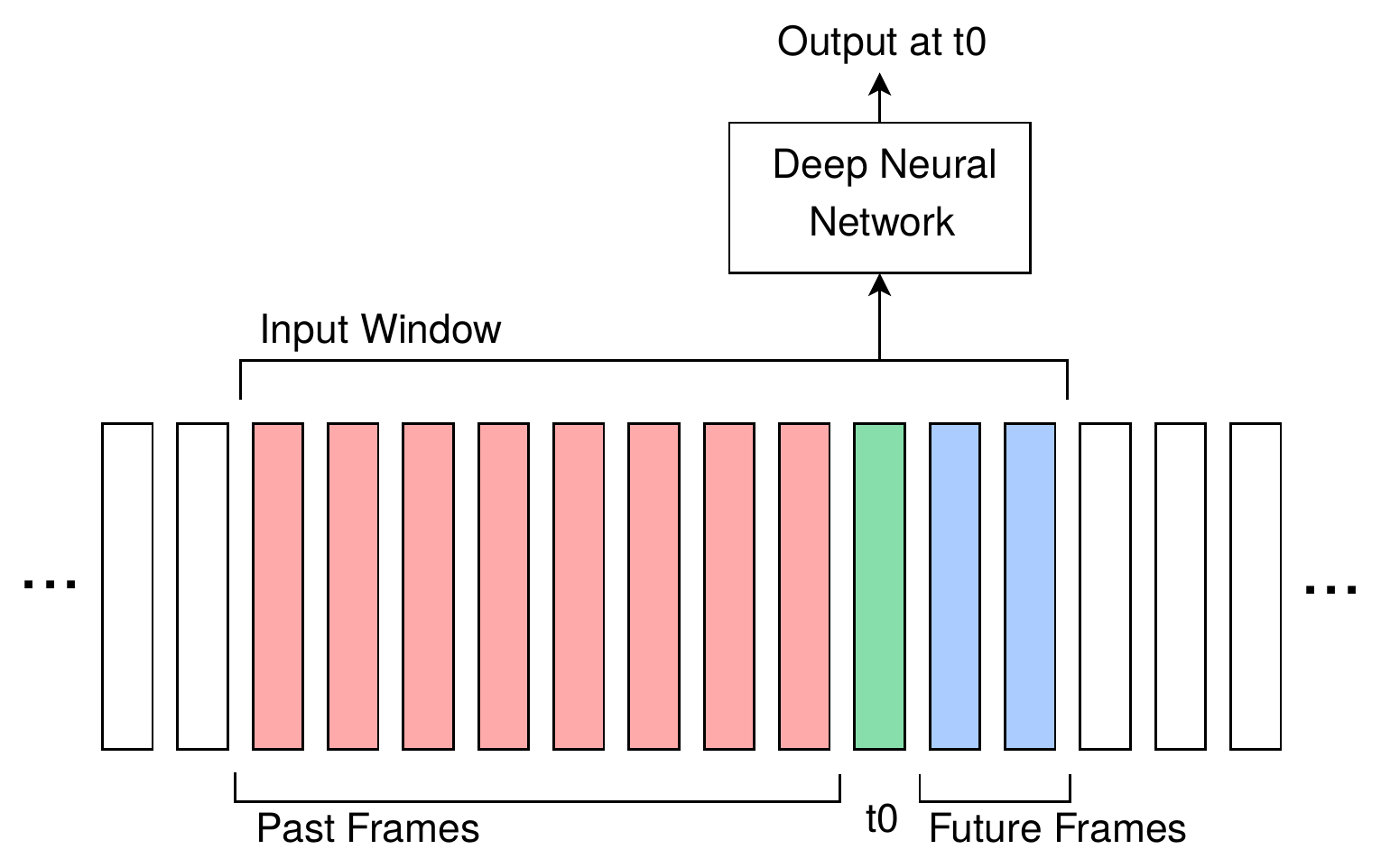}
\caption{Illustration of the method: A sequence of 11 speech feature frames constitutes the input to the neural network. The context window is not necessarily symmetric with respect to the current frame ($t_0$). In the illustration a shift of -3 was applied.}
\label{fig:dnnlatnn}
\end{figure}

Differently from previous studies \cite{Mohamed12, pdnn, dbn09, YaoEtAl2012SLTadaptation, DahlEtAl2011ICASSP}, we vary the alignment of the context window with respect to the current frame (see Figure~\ref{fig:dnnlatnn}). We train a different recogniser for each alignment and we analyse its performance in terms of Phoneme Error Rate (PER) as a function of the window shift.

\section{EXPERIMENTS}
\label{sec:experiments}
\begin{figure*}
\includegraphics[width=0.5\textwidth]{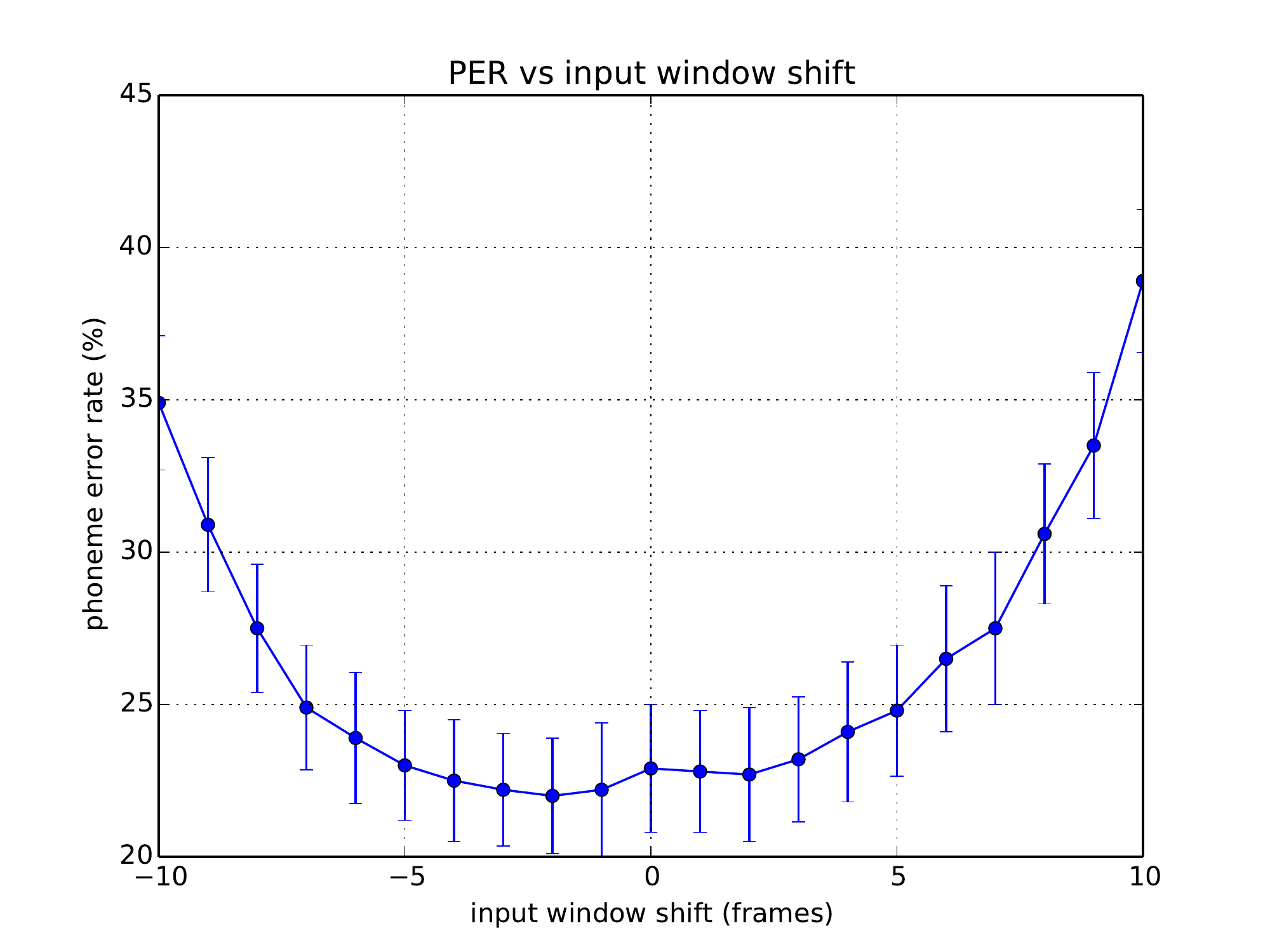}
\includegraphics[width=0.5\textwidth]{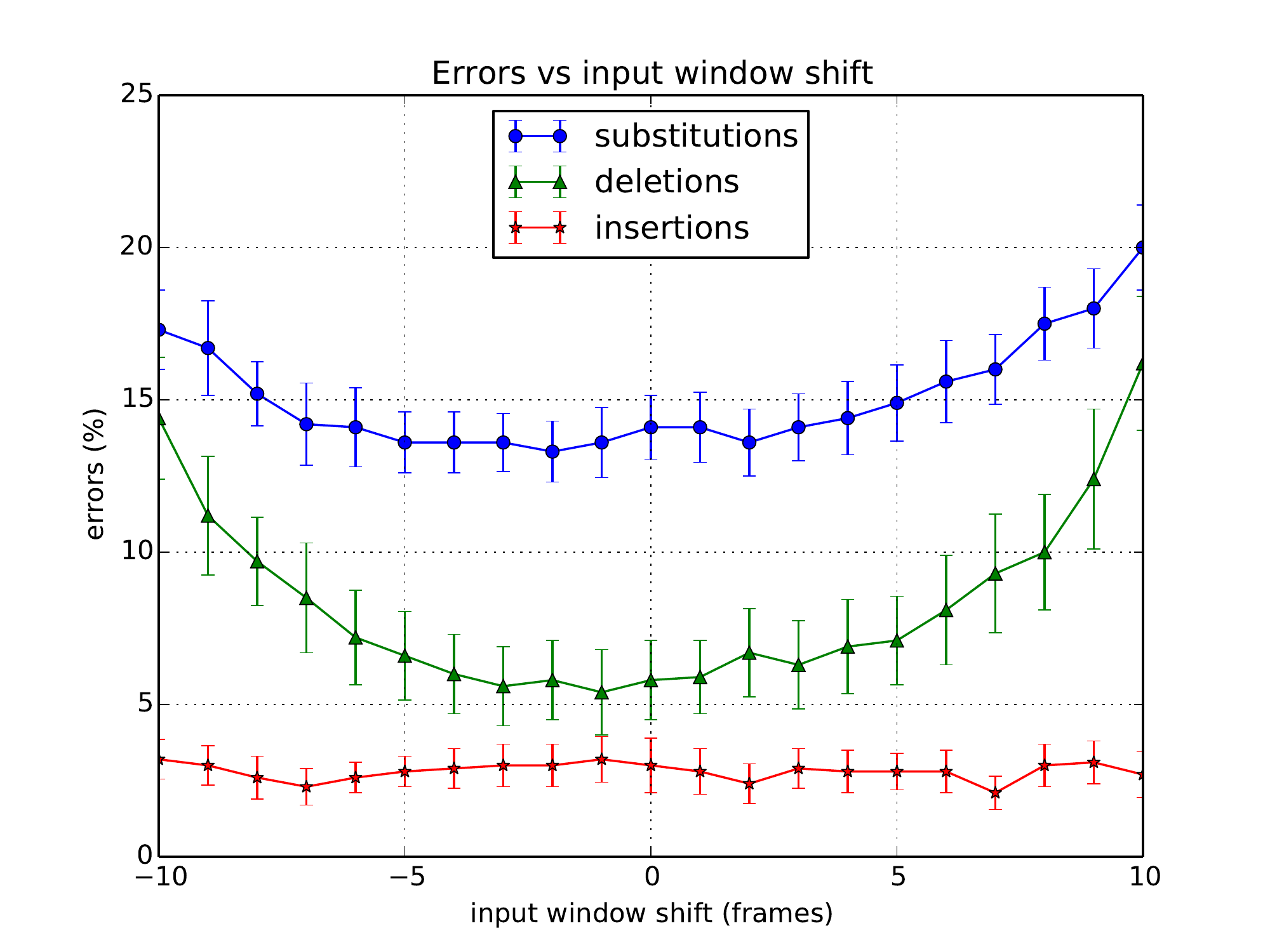}
\caption{Phoneme recognition performance as a function of the input window alignment in time. Left: phoneme error rate (PER). Right: detail on the different kind of errors. A shift of 0 corresponds to a window centered at the current frame with 5 frames right context and 5 left context. Windows with shifts above +5 or below -5 do not contain the current frame.}
\label{fig:results}
\end{figure*}

\subsection{Experimental Setup}
The experiments are based on the KALDI and PDNN+KALDI recipes \cite{PoveyEtAl2011ASRU-KALDI, pdnn} and are performed on the TIMIT corpus.
The standard 462 speaker training set was used for training.
All SA utterances were removed to prevent bias due to the similarity of the utterances.
The training set was further divided into 95\% training and 5\% validation set for regularisation during the back propagation training procedure.
Results are reported on the 24-speaker core test set.

The 40 channels filterbank features were computed using a 40~ms Hamming window with 10~ms increments.
The inputs to the DNNs in our experiments are context windows of 11 consecutive frames.
This length of context window seems to be optimal for this application according to \cite{Mohamed12}.
Also following the results in \cite{Mohamed12}, we use a DNN with 4 hidden layers of size 1024. The softmax output layer represents the distribution of posteriors over 1984 different senones (based on the 61 phonemes in the TIMIT standard phoneme set).
The resulting topology of the network is, therefore, $440\times 1024\times 1024\times 1024\times 1024\times 1984$.
The learning rate is initialised to 0.08, and then dropped by half whenever the difference between the previous epoch and current epoch drops below a certain threshold.

We optimised the acoustic scale in the decoder by running the decoder on the development set with 8 different scales.
The scales were in the form $1/k,\ k\in[1, \dots, 8]$.
The optimal values for the acoustic scale were always contained between the extreme values we tested, suggesting that they correspond to real optima, see also Table~\ref{tab:acousticscale} in Section~\ref{sec:results}.
These optimal values were then used to decode the test data.
The insertion penalty for our experiments was not optimised.
After decoding, the 61 phone classes are mapped to a set of 39 classes as in \cite{lee89} for evaluation. 

Our baseline results correspond to the input window being centered with respect to the current frame, with $5$ context frames on either side.
This corresponds, in our notation, to zero shift (see Figure~\ref{fig:results}).
A positive shift corresponds to a shift of context window in the future.
We tested shifts from $-10$ to $10$ with increments of one frame.
Additionally, we tested shifts of $-15$ and $-20$ frames.
For every shift value, the whole training and evaluation procedure is repeated.
It is interesting to notice that for shifts above $+5$ and below $-5$ the context window does not contain the current frame, and the recogniser will try to predict the current phoneme exclusively based on context.


\subsection{Practical Setup}
We used KALDI for feature extraction, selection of the senones and alignment of the senone transcriptions to the speech data.
To speed up the deep neural networks training, we used two NVIDIA TITAN GTX GPUs.
We also used the symbolic computations software Theano \cite{theano1}, which is well optimised to do symbolic algebra for GPUs.
The generative training of the RBMs took about 3 minutes for one epoch over the entire training set, and the fine-tuning with back propogation took about 2 minutes for one full pass.


\section{RESULTS}
\label{sec:results}
The results are shown in Figure~\ref{fig:results} for shifts between $-10$ and $+10$ and Table~\ref{tab:extremeresults} for selected shifts ($-20$, $-15$, $-10$, $-5$, $-2$, $0$).
The left plot in Figure~\ref{fig:results} displays Phoneme Error Rates (PER) as a function of input window shift, whereas the right plot details the different kind of errors (substitutions, deletions and insertions).

The best PER of 22.0\% (SD 3.8) occurs for a window shift of $-2$, i.e., for a window that is not symmetric around the current frame but has more past than future frames.

The performance does not degrade varying the shift up to $-5$ frames (the current feature vector is still included in the input window).
The PER, however, starts increasing when the shift is beyond $-5$ frames and the network does not receive the current feature vector in input.
If we consider positive shifts for completeness, we can observe a similar behaviour, although the graph is not perfectly symmetric and the PER for positive shifts is generally slightly higher than for negative shifts of the same amplitude.

The acoustic scale used for decoding was optimised on the development set and then used on the test set. As a comparison, Table~\ref{tab:acousticscale} shows the optimal values of the acoustic scale if optimised on the development and test set.
In most cases, the same optimal value was obtained.
In the cases when different values are obtained, the corresponding difference in \% PER was no grater than 0.5\%.

\begin{table}
  \begin{center}
    \begin{tabular}{ccc|ccc}
      \hline\hline
      window & \multicolumn{2}{c|}{acoustic score} & window & \multicolumn{2}{c}{acoustic score} \\
      shift  & test set & dev set & shift  & test set & dev set \\
      \hline
      -20 & 0.50 & 1.00 & 0 & 0.14 &   0.20 \\
      -15 & 0.33 & 0.50 & 1 & 0.17 &   0.20 \\
      -10 & 0.25 & 0.25 & 2 & 0.20 &   0.17 \\
      -9 & 0.25 &  0.25 & 3 & 0.25 &   0.20 \\
      -8 & 0.20 &  0.20 & 4 & 0.20 &   0.20 \\
      -7 & 0.20 &  0.20 & 5 & 0.20 &   0.20 \\
      -6 & 0.25 &  0.20 & 6 & 0.25 &   0.20 \\
      -5 & 0.17 &  0.20 & 7 & 0.25 &   0.20 \\
      -4 & 0.20 &  0.20 & 8 & 0.25 &   0.25 \\
      -3 & 0.17 &  0.20 & 9 & 0.25 &   0.25 \\
      -2 & 0.20 &  0.20 & 10 & 0.25 &  0.25 \\
      -1 & 0.20 &  0.20 & & & \\
      \hline\hline
    \end{tabular}
  \end{center}
  \caption{Acoustic scales optimised on the test and development set for each window shift.}
  \label{tab:acousticscale}
\end{table}

Looking at the right plot in Figure~\ref{fig:results}, we can observe that the number of insertions is relatively constant with respect to the window shift.
The substitutions increase when the window is shifted with respect to the current frame, but the errors that vary the most with window shifts are deletions.
\begin{table*}
  \begin{center}
    \begin{tabular}{ccccc}
      \hline\hline
      Shift & \% PER (SD) & \% SUB (SD) & \% DEL (SD) & \% INS (SD) \\
      \hline
      -20 & 77.6 (3.0) & 18.7 (2.4) & 57.9 (5.0) & 1.0 (0.7) \\
      -15 & 61.8 (4.0) & 21.6 (3.3) & 38.0 (6.1) & 2.2 (1.5) \\
      -10 & 34.9 (4.4) & 17.3 (2.6) & 14.4 (4.0) & 3.2 (1.3) \\
      -5  & 23.0 (3.6) & 13.6 (2.0) & 6.6 (2.9)  & 2.8 (1.0) \\
      -2  & \textbf{22.0} (3.8) & 13.3 (2.0) & 5.8 (2.6)  & 3.0 (1.4) \\
      0  & 22.7 (4.2) & 14.1 (2.1) & 5.8 (2.6)  & 3.0 (1.8) \\
      \hline\hline
    \end{tabular}
  \end{center}
  \caption{Recognition performance for selected window shifts}
  \label{tab:extremeresults}
\end{table*}
Table~\ref{tab:extremeresults} shows results for selected window shifts, including the extreme cases $-20$ and $-15$ that are not reported in Figure~\ref{fig:results} for clarity of illustration.
The performance for extreme shifts drops considerably and the degradation is mostly accounted for by deletions and substitutions.


\section{CONCLUSIONS}
\label{sec:conclusions}
This study presents a systematic analysis of the effect of shifting the context input window in a CD-DNN+HMM phonetic recogniser with respect to the current frame. The goal is investigating the possibility to reduce the latency of such a speech recogniser for applications with specific requirements, but the results reported here are of general interest.

Our results on the TIMIT database suggest that a context window slightly shifted back in time is superior compared to the symmetric context window used in most speech recognisers.
However, the improvement in performance is small compared to the variability (standard deviation), and this observation should be confirmed by testing on other data sets.

More interestingly, our results suggest that shifting the context window back in time up to $5$ frames (50 ms) does not introduce noticeable degradation in the system performance.
Larger shifts introduce a gradual but progressively steeper degradation.
As a consequence, without modifying the ASR method in \cite{pdnn}, we can reduce the latency of the system of at least 50 ms, without any degradation in performance.
We can reduce the latency even more if some degradation can be tolerated by the application.
This reduction in latency, although small in size, can potentially improve the usability of ASR in many applications, especially if latency is critical as in real-time lip synchronisation for telepresence.

It is important to note that the insertion penalty was not optimised in our experiments, and for all the different window shifts, the deletion error was always greater than insertion error.
In future work, we will investigate if we can reduce the effect of window shift by optimising the insertion penalty for each shift.

As in any study on speech recognition, the possibility to generalise our results outside the scope of phonetic recognition needs to be verified with specific tests. For example, it would be interesting to test if systems with longer time dependencies (lexical models and more complex language models), would be affected by the window shifts in a similar way.



\section{ACKNOWLEDGMENTS}
The GeForce GTX TITAN and TITAN X used for this research were donated by the NVIDIA Corporation. Giampiero Salvi is partially supported by the IGLU project (CHIST-ERA, Vetenskapsrådet 2015-06814). 

\label{sec:refs}

\bibliographystyle{IEEEbib}
\bibliography{strings,refs}

\end{document}